# Supporting peace negotiations in the Yemen war through machine learning

Miguel Arana-Catania[1]
Felix-Anselm van Lier[2]
Rob Procter[1*]

[1]University of Warwick and Alan Turing Institute for Data Science and AI, UK
[2]Government Outcomes Lab, Blavatnik School of Government, University of Oxford

*Corresponding author: rob.procter@warwick.ac.uk

**Abstract**
Today's conflicts are becoming increasingly complex, fluid and fragmented, often involving a host of national and international actors with multiple and often divergent interests. This development poses significant challenges for conflict mediation, as mediators struggle to make sense of conflict dynamics, such as the range of conflict parties and the evolution of their political positions, the distinction between relevant and less relevant actors in peace-making, or the identification of key conflict issues and their interdependence. International peace efforts appear ill-equipped to successfully address these challenges. While technology is already being experimented with and used in a range of conflict related fields, such as conflict predicting or information gathering, less attention has been given to how technology can contribute to conflict mediation. This case study contributes to emerging research on the use of state-of-the-art machine learning technologies and techniques in conflict mediation processes. Using dialogue transcripts from peace negotiations in Yemen, this study shows how machine-learning can effectively support mediating teams by providing them with tools for knowledge management, extraction and conflict analysis. Apart from illustrating the potential of machine learning tools in conflict mediation, the paper also emphasises the importance of interdisciplinary and participatory, co-creation methodology for the development of context-sensitive and targeted tools and to ensure meaningful and responsible implementation.

Policy Significance Statement
This study offers insights into how machine learning tools can be used to assist conflict mediators in organising and analysing data stemming from highly complex and dynamic conflict situations. Machine learning tools can bring significant efficiency to mediation by organising complex data and making it more easily accessible, giving mediators more control over existing information. They can also support consensus finding by highlighting areas in which political actors are converging or diverging; point to potentially overlooked areas of conflict or dialogue bottlenecks; and challenge prejudices that may have built up during a mediation process. This study shows how machine learning tools can bring about new innovative approaches for addressing mediation in increasingly complex, fluid, and protracted conflicts.

**Abbreviations**: NLP: Natural Language Processing; BERT: Bidirectional Encoder Representations from Transformers; LDA: Latent Dirichlet Allocation; NMF: Non-Negative Matrix Factorisation;

## 1. Introduction

In a recent report, the United Nations Secretary-General recognised the importance of technology for the UN's peace-making efforts, emphasising that 'engagement with new technologies is necessary for preserving the values of the UN Charter and the implementation of existing UN mandates' (United Nations Secretary-General, 2018, p. 4). While the broader field of peace technology is booming, data analytics technologies and techniques are still only used to "a lesser extent in the context of ongoing mediation efforts" (United Nations Department of Political and Peacebuilding Affairs & Centre for Humanitarian Dialogue, 2019, p. 12). While there is growing appetite among peace mediators for machine learning tools and international organisations are already deploying AI methodologies to inform broader peace operations, "it is yet to be determined what AI-based tools can contribute, substantially, to mediators' understanding of conflicts" (Lindström, 2020). This article uses first-hand data from peace dialogues in Yemen to contribute to emerging research in this field by exploring how Natural Language Processing tools can help mediators to make sense of dialogue dynamics.

We begin by making the case that machine learning-supported conflict analysis is becoming increasingly important for understanding today's dynamic, complex and protracted conflicts. We present a examples of recent research and practice on the use of digital technologies to support peace mediation. We find that while significant progress has been made in the analysis of contextual conflict data that surrounds peace mediations, very little research has addressed the question of how machine learning tools inform the analysis of peace mediations themselves. We then turn to our case study, where we describe the challenges a team of international mediators faced in facilitating peace negotiations in Yemen, particularly in keeping track of long-drawn out and dynamic peace negotiations. We explain the nature of the data made available to us and the development process of machine-learning tools, which built on a participatory design and co-creation methodology.

We conclude with a discussion on different use cases that were drawn from feedback received from the mediation team. We explain how the tools have the potential to (1) facilitate data extraction and management, which eases the mediators' access to relevant data; and (2) support mediation process management by helping mediators to better grasp dialogue dynamics and party positions. Finally, we emphasise the importance of a participatory, co-creation methodology for the development of context-sensitive and targeted tools and to ensure meaningful and responsible implementation, and we highlight some of their limitations.

## 2. Complex and protracted conflict scenarios

As the nature of armed conflicts and their trajectories has been transforming over the last 30 years, international conflict mediation approaches are constantly being adapted and the mediators' skillsets expanded (da Rocha 2019). In particular, international mediation organisations are increasingly exploring how digital technology may support peace making efforts to deal with the new complexity of conflicts as well as with the increasing amounts of data that are produced in and around peace processes (Hirblinger 2020a).

While the number of traditional symmetrical conflicts fought between states is declining (i.e., between armies), there is an increase in intrastate violence and asymmetric wars (i.e., between state armies and non-state actors), including civil wars, insurgencies, terrorism, guerrilla wars and large-scale protest and violence. Conflicts tend to be more fluid and fragmented, involve increasingly complex webs of state and non-state actors, and often spread across borders and affect broader regions (Strand et al., 2019). As conflicts and their environments have become more dynamic, complex and protracted, so have the mediation processes that try to solve them. Peace operations today tend to be increasingly drawn out and complex processes, making it difficult for mediators to keep track of conflict developments (Brahimi & Ahmed 2008; da Rocha 2019).

These new empirical realities pose significant challenges to the international community's peace efforts. Traditional forms of international peace interventions appear increasingly ill-equipped to address current conflict environments (Avis, 2019, p. 4). In the context of heightened conflict dynamics and complexity, mediators struggle to generate a solid understanding of how a conflict's context affects dialogue dynamics, including the range of relevant conflict parties and the development of their political positions, or the identification of key conflict areas and their interdependence. In order to fulfil their role effectively "peacemakers must integrate a more sophisticated analysis of technological factors into their broader analysis and engagement strategies" (Kavanagh 2021).

## 3. Peace making and machine learning

While for a long period of time, work in this field remained theoretical, researchers and practitioners are increasingly exploring how AI-tools can effectively support conflict mediation in practice. For example, in 2018 the UN Department of Political Affairs launched a "CyberMediation Initiative", together with other leading mediation organisations such as swisspeace, the Centre for Humanitarian Dialogue and the DiploFoundation, to explore "how digital technology is impacting the work of mediators in preventing and resolving violent conflicts worldwide" (Centre for Humanitarian Dialogue, 2018). Several recent reports and studies offer overviews of different scenarios in which technology can support mediation efforts (Höne, 2019; Jenny et al., 2018; United Nations Department of Political and Peacebuilding Affairs & Centre for Humanitarian Dialogue, 2019).

Several research projects focus on developing tools that address the problem of dynamic societal conflict dynamics in which mediation processes take place. UN agencies now regularly apply machine-learning based analysis on information sourced from social media networks and traditional news media to determine social and cultural attitudes, intentions, and behaviour (Pauwels 2020). The UN Innovation Cell, for example, used data mining techniques and AI technology to identify and analyse the contributions of influencers, fake news and trending topics on social media in Somalia; it has also analysed peoples' voices from radio broadcasts to identify (and potentially prevent) violence, conflict and social tensions (UN Innovation Cell 2018). Such tools have the potential to support mediators by making them aware of the attitudes of particular societal groups towards the conflict and pointing them to potential shifts of the mediation context.

As modern conflicts have significantly broadened the set of conflict actors, other researchers emphasise the need to use digital technologies to increase the inclusivity of peace processes. Hirblinger (2020a) notes that digital inclusion has the potential to increase the legitimacy of peace processes and their outcomes and reduce the risk of continued violence by involving actors beyond the main parties to the conflict, empowering marginalised groups, and transforming conflict relationships by focusing on relationships between conflict parties and broader sets of stakeholders. A practical example of how aspects of such digital inclusion may be supported with AI-enabled technology is a recent collaboration between the UN and Remesh, a US-based start-up specialising in processing large-scale online conversations in real-time. The UN trialled the tool in Libya, where it helped UN officials to engage with a broader set of people, channel their voices into their negotiations, and ensure both the transparency and credibility of the process (Brown, 2021; Warrell, 2021).

Fewer studies, however, have looked at how machine learning can help make sense of the "inner workings" of peace mediation, that is, the structured negotiations between a set of conflict parties aimed at the prevention, the management or resolution of a conflict. As mediation processes are becoming increasingly protracted and complex, it is becoming increasingly difficult for mediators to keep track of multi-year and multi-track negotiations with multiple stakeholders and their evolving negotiation standpoints. One promising avenue of fostering a better understanding of internal dynamics of peace negotiations is argument mining and argument analysis. These techniques seek to automatically extract structured arguments from unstructured textual documents (Lippi and Torrino 2016; Betz and Richardson 2021). The "Mediating Machines" project team, set up to explore the potential of AI in mediation processes, investigated avenues for translating the advances of AI-enabled argument analysis techniques to better understand the opinions voiced by conflict parties and stakeholders (Hirblinger 2020b). While these types of analyses have the potential to contribute to an understanding of a conflict matter by systematically dissecting the reasoning of conflict parties and shedding light on argumentative logics in stakeholders' reasoning, they have yet to be tested on first-hand peace dialogue data.

While significant progress has been made in technology and mediation, there is still "a lack of concrete examples and discussions that could bring discussions forward" (Mediating Machines, 2020). This is partly likely to be the case because of the limited availability of peace dialogue data - if recorded at all, data of peace negotiations is usually kept confidential. Accordingly, as of now, there are very few studies that have attempted to analyse real-world data of ongoing peace negotiations through machine learning tools. We hope that this study will highlight the potential benefits of machine learning-assisted analysis for peace mediation and encourage mediation missions to collect data more systematically and allow for further sandboxed experimentation.

## 4. Case study context

Up until the conclusion of a (still fragile) peace deal in early 2022, Yemen had been subject to a protracted regionalised war since 2014. The war comprised multiple armed and political conflicts, which had eroded central government institutions and fragmented the nation into several power centres. The roots of the war stemmed from a failed political transition that was supposed to bring stability in the aftermath of an Arab Spring uprising that brought down Yemen's long-time president Ali Abdullah Saleh.

RESEARCH ARTICLE
Supporting peace negotiations in the Yemen war through machine learning

From the outbreak of the civil war, the United Nations, the Office of the Special Envoy of the Secretary-General for Yemen (OSESGY) and other international actors had launched multiple attempts to facilitate talks between conflict parties to reach political agreement on the conflict issues (Palik & Rustad, 2019). Despite these efforts, the conflict continued to grow significantly in complexity. While the conflict was often portrayed as a conflict between two main forces, these two broad coalitions were becoming increasingly fractured and loyalties are fluid. A multitude of actors, both national and international, engaged directly or indirectly in the conflict, motivated by various divergent goals. Years in to the conflict, the political and military situation on the ground remained highly dynamic, with increasingly intransigent and divided parties and continuously shifting party positions (International Crisis Group, 2020).

This context complicated the mediator's goal to find consensus and agreement for political settlement. As party positions were volatile and increasingly difficult to track, the perceived effectiveness of dialogue efforts suffered and it became increasingly unclear whether any progress was made. This context also affected the work of the organisation for which this study was done. The organisation had been providing support to OSESGY's efforts to reach a peace agreement by acting as mediators in Track 2 and Track 1.5 peace negotiations since 2016. The primary goal of this study was to find ways in which machine learning tools could make the data collected over the years of dialogue more accessible and navigable to the mediators and to find new ways to assist them in the analysis of the data. This included, for example, finding ways to help the mediators to identify how party positions had evolved over time, i.e., whether parties had moved closer towards a consensus on particular dialogue issues or not.

## 5. Methodology overview

Knowing and understanding the social and political context of their application is vital for developing digital tools that are both effective and ethical. This is particularly important when machine learning tools are employed on highly sensitive subject matter such as conflict mediation. The project was designed in an interdisciplinary manner, involving both data scientists and socio-legal scholars. This approach was applied throughout the project, starting with the in-depth, empirical analysis of the case at hand, which provided contextual information informing the development of machine-based analysis of the parties' dialogues; to the development of machine-learning tools and the interpretation of the results they produced; to considerations about how data outcomes could be presented in a meaningful way to mediators.

The project followed a participatory design, co-creation methodology (Bodker et al., 1995). Since its emergence in the 1980s, this has become acknowledged as being key to successful IT projects (Voss et al., 2009). Arguably, following a participatory design, co-creation methodology has become especially important as machine learning techniques that are still so unfamiliar to many are now being widely applied to new products and services (Slota, 2020; Wolf, 2020). This unfamiliarity can lead to unrealistic expectations of their capabilities (Tohka & van Gils, 2021). Project progress was discussed in bi-weekly meetings with the conflict mediation team. Initially, these focused on building common ground between the participants: (a) familiarising the mediators with natural language processing (NLP) and machine learning concepts, techniques, capabilities and limitations; (b) enabling us to gain an understanding of the dialogue process, the data that it generated and how; and (c) establishing some initial requirements for how machine learning tools could support the dialogue process. As the project progressed, we were able to present a series of prototypes to the mediation team for their comments and feedback. This enabled the requirements of the tools to be progressively refined around an evolving set of agreed use cases that captured more specific ways in which the tools could be applied in support of the dialogue process and conflict mediation.

Computational Grounded Theory is a rapidly developing field and several methods have now emerged (e.g., Nelson, 2017). However, from our initial discussions with the mediation team, it became clear that it would be important not to try to follow any particular method but to let the method – and hence the tools – to be driven by our discussions with them. This eventually led to the development of: (1) two information extraction tools to organise dialogue text into predefined categories and to derive latent issues from the dialogues text; and (2) data analysis tools to measure party distances (i.e, how close or how far apart the positions of the parties are) on specific topics. Although each of the tools independently provides particular insights into the dialogue data, more holistic and meaningful insights into dialogue dynamics can be extracted when they are used in combination. In the following, we will describe how the tools were developed but also how the results of each tool need to be triangulated with other sources of data, both qualitative and quantitative, to validate the output and reach meaningful conclusions. Technical details of the methodologies used to develop the tools are described in sections 6, 7 and 8.

In addition to the regular meetings with the mediating team throughout the project, we conducted a feedback session at the end in order to get the mediating team's views on what the project had achieved and how the tools had supported their work. In particular, the sessions were designed to understand the weaknesses of the tools and how they could be further developed and improved. As peace negotiations had been suspended around the time the project ended, the mediation team had not yet had an opportunity to use the tools on new data. Thus our discussions focused on the analysis of data from the preceding 2 years of negotiations.

## 6. Data

The data for the project was mainly sourced from the dialogue sessions and expert meetings between Yemen's key stakeholders that the mediation team had conducted between 2018 and 2019. Each dialogue session was documented through "rough notes". Each note covered working sessions held between 2018 (6 sessions) and 2019 (8 sessions) and contains around 12,000 words for each session. All the notes together add up to 177,789 words. These rough notes served as the basis for the dataset which was used to develop the machine-learning tools.

These notes are non-verbatim transcripts of dialogue sessions, excluding all unnecessary speech without editing or changing the meaning or structure of the dialogue between parties. This means that the notes do not contain any "thinking noises" such as, "um", "uh", "er," "ah," "like," or any other utterances that would indicate a participant's feelings. However, they still include extensive detail of the substance and structure of dialogue sessions: the notes maintain the sequential and temporal structure of the discussions; record the participation of each representative present; and report the key arguments made by each representative every time they spoke. Below is an anonymised extract from the notes:

> - Party Rep. 1: problem is not the form. After 1990 we tried joint presidential council, failed directly after agreement. Tried advisory council, didn't work very well. Presidential council in Sanaa – agreed that President would rotate but hasn't. Problem is agreement on powers – have to detail clear powers of each institution and position. Avoid giving excessive powers to these executive bodies. Give limited powers. Many should be transferred to local and provincial levels, e.g. reconstruction, security. Reduce pressure on central government and reduce power grabbing. Whether in presidential council or in government.
>   - For a ceremonial president. Main powers should be with government. Or power sharing in presidency and government is purely executive.
>   - Should split up decision-making powers, not all in one body or person.
>   - Inside government, should have core bloc of ministers would take decisions. Strategic decisions require higher majority, others simple majority.
> - Party Rep. 2: we need to discuss options that are possible. We have frameworks we must not ignore. We were against presidential council because contradicts frameworks, and goes with constitutional declaration and coup. History of Yemen since revolution – every presidential council formed failed, followed by war. For Yemen to be led by several people, all president, won't work and violates frameworks. Yes there is a problem in presidency and monopolisation. How to reduce this and reform the institution. E.g. activate the council of advisors. What are mechanisms for decision-making? Or could have Vice Presidents with specific dossiers and decision-making powers. Otherwise could destroy everything we agreed on.
> - Party Rep. 3: any model can succeed in one context and fail in another. Collective and individual presidency – not absolutely good or bad. In Yemen, before Ali Abdullah Saleh, there were presidential councils. In south under unification, there was presidential council and semi-parliamentary system.
>   - Last amendment of constitution was in 2009 – art 65 re parliamentary term.
>   - Need to bear in mind current constitution until a new one is adopted. Yes may be amended by future agreement but not totally repealed. GCC Initiative amended some parts but did not repeal.

The notes were taken by a notetaker from the mediating team, which is bound to offer its mediation support independent of specific national or political interests. Finally, the notes were taken in Arabic and subsequently translated into English by a professional translator. While non-verbatim note taking and subsequent translation may introduce some errors and bias, the source of the data and the data collection process still instils a reasonable amount of confidence that the notes capture the salient features of the deliberations between the different conflict parties.[1]

In addition to this dataset we also had access to internal documents, including over 30 detailed meeting reports, as well as the organisation's own systematic "comprehensive analysis" of dialogue developments. These documents provided important contextual information that helped guide and structure the development of the tools and triangulate the results of the data analysis.

## 6.1. Data preparation tool and pre-processing methodology

Before any tool development for the analysis of this data could be undertaken, the notes had to be cleaned and pre-processed and transformed into a structured dataset for the analysis. To this end, we produced a tool for data pre-processing and cleaning, which is capable of processing both existing data and any future notes provided they are formatted in the same way.[2]

The pre-processing of the texts includes: deleting non informative, Unicode characters from the original word documents; identifying indentation format of the dialogues in order to build the conversation threads correctly; deleting non conversational text; correcting and uniformising entities spelling; detecting text shared by several entities; and abbreviation expansion. The pre-processing is done with the help of the NLTK[3] package and the Pandas[4] library.

The tool then automatically extracts and organises data from the notes into an CSV file, organising the dialogues comments in the following fields: text, original raw file name, year, month, participant name, participant organisation, and participant multi-organisation (in case several parties are sharing a statement).

This dataset provided the basis for filtering and extracting more nuanced information via machine learning tools at subsequent stages of the process. In addition, the Data Preparation Tool generated a CSV file that can be used by the mediators to carry out a simple information analysis or to retrieve specific parts of the dialogues. For example, it is possible to obtain the comments made by a specific party or a subset of parties within a certain time interval.

## 7. Issue extraction tools

Following initial discussions with the mediation team, it was determined that the basic requirements that of these tools should satisfy would be to: (1) categorise the dialogue texts into a set of issues predefined by the mediators; and (2) identify latent issues that emerged from the dialogue text without manually pre-defining them. To do so, we proceeded with two different approaches: query-driven topic modelling and topic modelling.

### 7.1. Predefined issue extraction: query driven topic modelling

After years of facilitating dialogue, the mediators wanted to systematise all available notes and recordings into a "comprehensive analysis" of the dialogue sessions. This "comprehensive analysis", which entailed a list of what the mediator perceived as key conflict issues, was to provide an overview of dialogue development, as well as to organise the dialogue sessions and to sketch out avenues on how to reach agreement on these issues. This list of issues is important as it reflects the organisers' first-hand knowledge of the dialogues, as well as their own particular vision and analysis of the situation. However, the drafting of the comprehensive analysis proved challenging as the mediation team had to manually extract information from the available data, which was both ineffective and costly. The goal of the predefined issue extraction tool was to match dialogue text to 18 predefined conflict categories or issues defined by the mediation team in their own analysis.

---

1 The research team was given access to this data for internal analytical purposes. However, given the high-stakes environment of the dialogue sessions, the sensitivity of the topics discussed, and the potential influence of the release of the data on ongoing peace negotiations, we have not been granted permission to release the data publicly.

2 We have produced a style guide to ensure that any future notes can be read and analysed and any automatic text analysis tool is able to extract as much information as possible in the most efficient way.

3 https://www.nltk.org/

4 https://pandas.pydata.org/

### 7.1.1. Method

To create the predefined issue extraction tool, we used techniques based on query-driven topic modelling, where keywords relevant to each issue are predefined. The tool uses word representations produced with a model trained on a large corpus of text to understand how words are used in relation to each other, and be able to detect words with similar meanings or relating to the same matters. In this case the technique uses this knowledge to detect words in the comments that are related to the predefined keywords for each issue. For example, for the query term "State Institutions", the tool will not only identify passages in the text which contain the exact word but also related words such as "ministry, district, government", etc., because these have similar meanings to the query term.

The procedure consisted of defining query words for each of the 18 issues and detecting in the dialogues terms whose vector representations (embeddings) were found within a certain distance of the queries. The comments of the dialogues containing the original keywords or near terms were categorised as belonging to the issue. It was possible to categorise a text with multiple issues. Term definitions and the qualitative analysis of the results were conducted in collaboration with the mediation team.

We first defined an initial set of keywords for each issue, and compared the near terms found in the texts by using two general types of non-contextual word representations, Word2Vec (Mikolov et al., 2013a, 2013b) embeddings produced by Google and GloVe embeddings (Pennington et al., 2014) produced by the NLP group at Stanford University[5]. The GloVe embeddings were found to produce more meaningful results. For the rest of the evaluation we used the pre-trained GloVe embeddings "glove.6b", using 6 billion tokens represented in a 300 dimensional space. The embeddings are trained in the Wikipedia 2014 dump and in the English Gigaword Fifth Edition corpora (Parker et al., 2011). This selection was also done considering the need of the tools to be lightweight enough to run on commonly available computers by the mediation team. Stop words from the texts were removed by using the NLTK package. No further pre-processing steps were performed beyond those described in section 4.1. The distance between terms was computed as the cosine similarity between the embeddings by using the Gensim[6] library (Rehurek and Sojka, 2010). The threshold distance was defined dynamically, starting from a common value but with the possibility of reducing the distance for each term in case of obtaining too many related words.

The fine tuning of the tool included, as a first step, adjustment of the parameters and threshold to identify related terms in the query. The only parameters involved in this technique are the similarity range and maximum number of words to consider. After examining the outcomes of other configurations we selected a minimum threshold for the cosine similarity of 0.4, with the possibility of increasing it up to 0.6 when finding more than 1000 similar words. Next, we concentrated on refining search terms. The search terms for each issue needed to be iteratively refined for the tool to be able to distinguish between and associate text relevant to different issues. For example, to identify relevant text for the issue "The South", search terms such as "reparations", "independence", and "autonomy" were used.

Once the list of search terms for each issue was consolidated, we further refined the search by experimenting with different types of queries. First, we tested the results when the tool performed a search for each term individually, which means that the tool would associate text identified by an individual search term with a particular issue. We also tested the results when all the search terms were combined. This final combination of searches with all keywords for each issue was used to produce the final categorisation of comments into the predefined issues. The system produced as an output not only the list of comments categorised, but also the query expansion for each of the initial queries of each category with the new near terms found for any future refinement of it.

The methods were applied iteratively, producing the first results using generic parameters and a first set of keywords. These parameters and keywords were then refined together with the mediation team at each iteration.

---

5         https://nlp.stanford.edu/projects/glove/
6         https://radimrehurek.com/gensim

Hence, the final parameter settings and set of keywords is the result of a series of internal evaluations conducted collaboratively with the mediation team. We evaluated the automatic query expansion of the terms, which offers a global view of the keywords detected in the conversations, and the specific text categorised in each issue.

The validity of the outcomes was further substantiated by means of a final evaluation and review of that data by the mediators. This final evaluation confirmed that the topic modelling tool was capable of adequately categorising text into predefined issues, thereby significantly reducing the workload of the mediation team which has previously executed the same categorisation of text manually.

### 7.1.2. Discussion

The results of the machine-learning analysis were presented in a CSV file, which categorises each comment made by each participant into one (or more) predefined categories. The filtering function of any spreadsheet software allows mediators to filter comments according to different parameters, including a particular dialogue category but also according to a year or month in which comments have been made, and particular parties or dialogue participants involved. Such a tool allows for the effective navigation and systematic exploration of dialogue data and may prove to be useful, for example, to hold participating dialogue parties accountable for their positions.

The tool also allows for a meta-level analysis of the data. In addition to categorising pieces of text, the prominence of individual issues were measured by the number of words, which allows for insights into broader dialogue dynamics. By looking at the words-per-issue graph (Figure 1), we can identify the amount of debate that each issue generated over the course of a year and how this changes over time. The data shows that there has been a general increase of dialogue activity from 2018 to 2019, as the number of words per issue approximately doubled from one year to the next. Looking more closely into each issue, we can also identify how intensively individual issues were discussed over time. Here, we can observe that while some issues remain relatively stable in terms of how intensively they are being discussed, others tend to fluctuate. For example, while discussions concerning "Decentralisation/federalism", "Dispute resolution" and "National body" increased from 2018 to 2019, the issue of "Government of National Unity" – and to a lesser extent "Demobilisation", "Guarantees", and "Sequencing of negotiations" saw a decrease of discussion.

These 'number of words' graphs merely represent a quantitative analysis of the volume of text associated with a particular topic. They are not necessarily indicative of the substantive relevance of each of the topics discussed to the mediation process. Nevertheless, the graphs provide an additional source of information about the dialogues, which may support the mediators' analysis of the mediation process.

For the purposes of interpretation of the data, it is important to remember that comments are categorised into one category or another based on their word content. This means that the activity reflects what has been said, regardless of whether the debate was organised at the time to address that issue. Thus, a topic may be debate-heavy because of the mediator's thematic emphasis in the dialogue sessions, but they may also highlight the issues that produced the most engaged discussions among the participants.

## 7.2. Latent issue extraction: topic modelling

The second type of issue extraction was designed to automatically extract, identify and classify the most relevant issues raised by the participants throughout the dialogues. Rather structuring the dialogues into a set of predefined issues, the tool identifies the most relevant issues as they emerge organically from the discussions, based exclusively on the textual content of the dialogues. This second method of issue extraction offers a new perspective on dialogues' substantive focus and may point mediators to aspects they have not yet considered.

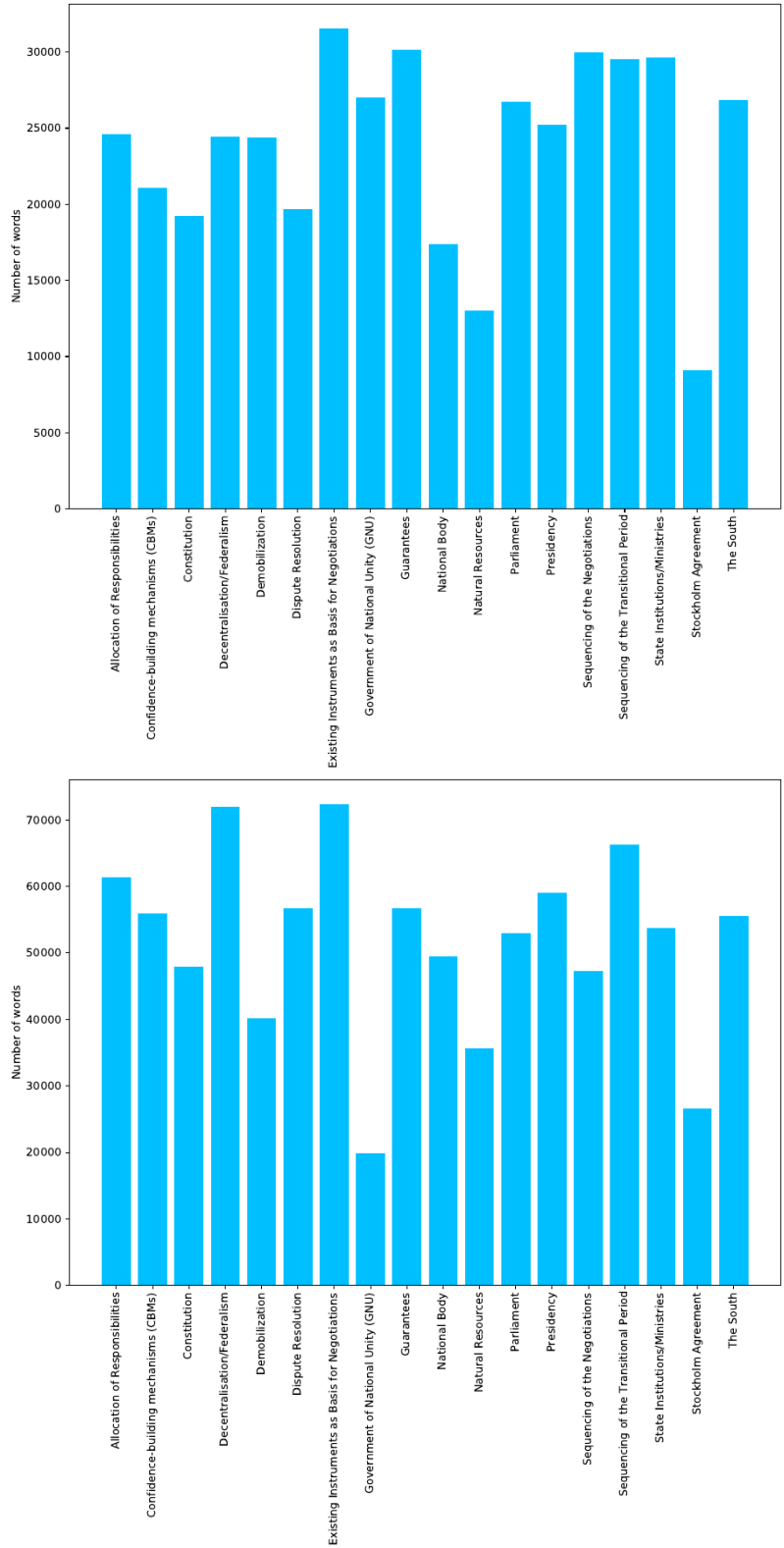

*Figure 1: Relevance of predefined issue by number of words for 2018 (top) and 2019 (bottom).*

### 7.2.1. Methods

Two different topic modelling techniques, Latent Dirichlet Allocation (LDA) (Blei et al., 2003; Hoffman et al., 2010) and Non-negative Matrix Factorization (NMF) (Hoyer 2004; Lee and Seung 2001; Paatero and Tapper 1994) were tested and evaluated for the latent issue extraction. The underlying idea of both techniques is the same: given a set of documents in a corpus (in this case we define each comment made by a participant to be a distinct document in the dialogues corpus), topics consist of a set of words and each document is a mix in some proportion of one or more topics. NMF, iteratively searches for a matrix decomposition of the original matrix of documents into words. LDA, as described in its references, understands the documents as generated by probabilistic distributions over topics and words, where, in particular, the probability of each topic in each document is given by a multinomial Dirichlet distribution.

LDA and NMF were experimented with to automatically extract between 5 to 30 issues, each of which was then defined by 10 keywords that, based on their frequency of occurrence, were the most representative of the words that make up that issue. Together with the mediation team we then analysed the meaningfulness of the results by scrutinising the keywords of each issue as well as the top 10 extracted participant comments for each issue.

Following evaluation, as will be explained at the end of this section, it was determined that NMF produced better results. In particular we used the NMF implementation of the scikit-learn[7] package (Pedregosa et al., 2011), using the Frobenius Norm as the objective function, a Term-Frequency Inverse Document-Frequency (TF-IDF) representation of the words, Non-negative Double Singular Value Decomposition (Belford et al., 2011; Boutsidis and Gallopoulos 2008) for the initialization and a coordinate descent solver using Fast Hierarchical Alternating Least Squares (Cichocki and Phan 2009; Gillis 2014). Stop words from the texts were removed by using the NLTK package. For the vocabulary used in the topic modelling we considered the top 10000 features and excluded terms appearing in more than 90% of documents. Regarding the parameters of the NMF decomposition, we used a regularisation parameter alpha with a value of 0.1, and a regularisation ratio of 0.5 for the mixing between the L1 and L2 regularisation, and a tolerance of 1e-4 to the stopping condition. NMF outputs directly the decomposition of documents in each of the possible topics, without the need to identify the keywords in the documents and then identify the keywords in the topics. A 0.1 threshold was used as a percentage threshold to classify a document as belonging to a topic.

We then refined the topic modelling tool. For example, we tested whether the tool would perform better if it only considered nouns in the dataset. Further, we noticed that the structure of "working sessions" outcomes (i.e., multiple parties presenting results of group discussions on particular issues) tended to obscure the topic modelling process, as the substance of this text was often more technical in nature, making it more challenging for the tool to classify text meaningfully. We decided to drop multiple party responses from the overall analysis, which then produced the most well-defined issues and provided the richest forms of information. No other additional pre-processing steps were applied besides the ones mentioned here and in section 4.1.

As before, the evaluation was carried out in collaboration with the mediation team. Each of the methods, as well as the parameters used, were analysed by evaluating the results produced. These topic modelling techniques allow not only a detailed evaluation of each result, but additionally rank each element in terms of its match to each topic. In this way, the most representative texts of each topic can be obtained in a simple way, which offers a global evaluation of the techniques used, complementary to the exhaustive evaluation of each text. During the feedback session, the mediation lead confirmed that the results of the data "generally matched how we understood things to be at the time." (Feedback Session, 2022).

As we have pointed out in previous sections, this methodology allows a global and simultaneous analysis of all the mediation sessions, which is unfeasible to be carried out manually. In this way, a unique additional point of view is offered to the mediation process.

### 7.2.2. Discussion

The latent issue extraction tool allows for a more unconstrained analysis of the issues that emerge naturally from the dialogues. This approach may highlight aspects of the negotiations that appear to be of particular importance to the dialogue participants and may offer new perspectives on dialogue dynamics. While the distribution of the topics may still broadly reflect the mediator's choice of dialogue structure, the data reflects what participants actually said during the dialogues and how their interventions shaped what was being talked about at each moment. When comparing the issues generated through latent issue extraction with those predefined by the mediator we found some overlap. For example, questions about natural resources, the composition of the national body, the South, or sequencing were issue categories that appeared in the results of both extraction tools. However, the latent issue extraction tool also brought to light several "new" issues, such as representation, disarmament, the role of the UN, etc. While each of these issues may be categorised into one of the issues previously identified by the mediator, they nevertheless highlight aspects of the negotiations that appear to be of particular importance to the participants. This information could be used to reconsider the substantive focus of future dialogue sessions.

---

7 https://scikit-learn.org

The tool produced a list of new issues and organised all relevant text into those issues (see Table 1 and Box 1 for an explanation).

| Issue | Issue Description | Keywords |
|---|---|---|
| issue 0 | Sequencing | political forces regarding military conflict security want war yemen Sanaa |
| Issue 1 | Executive Powers/Composition | president powers vp presidential_council vps presidency advisory_council prime_minister decisions legitimacy |
| Issue 2 | Institutional Security Arrangements | would military_security_committee committee formed option implement third_party independent two monitor |
| Issue 3 | Framework Agreements | agreement hodeida implemented sides implementation framework interpretation_committee implement signing stockholm_agreement |
| Issue 4 | Implementation Issues | need address new solution violations realistic transition take end_war participation |
| Issue 5 | Sanctions | sanctions said guarantees yemeni talking actors think implementation agreements |

*Table 1: First 5 latent issues and top ten keywords for 2019.*

| Issue | Issue Description | Keywords |
|---|---|---|
| Issue 0 | Representation/Appointments (e.g., technocratic/political) | political parties minister social competence arms achieve important real conditions |
| Issue 1 | Allocation of Responsibilities | government president presidency political_forces gcc_initiative parliament technocratic would forces decision_making |
| Issue 2 | Natural Resources | resources regions federal natural draft local revenues given model draft_constitution |
| Issue 3 | National Body | national_body committee composition guarantees implementation current oversee body disputes role |
| Issue 4 | South | south make issue southern since represent southern_issue groups yemen international |
| Issue 5 | Disarmament/Guarantees | state legitimacy arms even armed militias groups issue solution |

*Table 2: First 5 latent issues and top ten keywords for 2018.*

> Each issue is defined by the most frequent words that appear in the comments regarding the issue. These keywords are presented on the right side of Table 1.
>
> Each comment can be tagged as belonging to several issues at the same time (e.g., somebody talking about "Sanctions" and "Agreements" in the same paragraph). In order to understand the issues better, the tool also extracts the top 10 most representative comments for each issue. That is, the comments that are more uniquely identified by the specific issue in comparison with others.
>
> The list of representative comments, together with the list of keywords of each issue, help to better understand what is being discussed. Using this information, we produced a manual description of each issue, which can be seen on the left side of the previous tables.
>
> *Box 1: Explanation of Tables 1 and 2.*

This form of issue generation can be used in various ways. Most importantly, latent issues offer an alternative perspective of the way the dialogues unfolded in practice. The relevance of issues by number of words graph (see Figure 2), shows the activity of parties in each of the latent issues. For example, the data indicates that questions surrounding the issue number 0 "sequencing" appear to have been discussed intensively in 2019. The disproportionate amount of text in this issue is also indicative of significant overlap of the broader issue 0 "sequencing" with other issues. This was confirmed by measuring the overlap between issues. As we pointed out above, these 'number of words' graphs should not be understood as direct reflections of the importance of each issue. The quantitative data obtained should be taken into account by the mediation team as an additional source of information to understand the evolution of the discussions and to plan future steps.

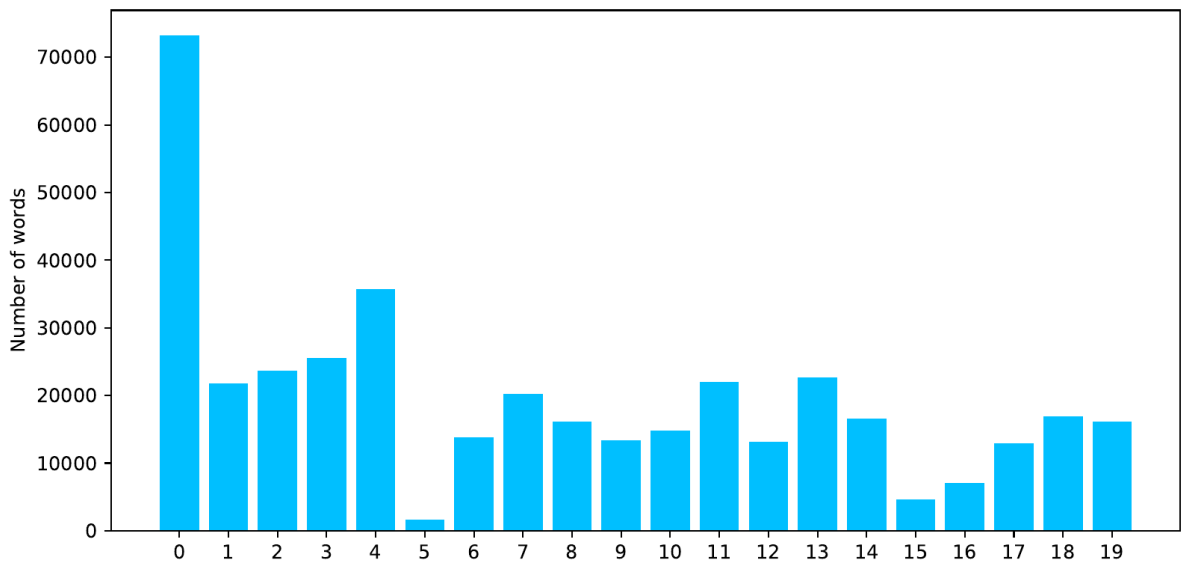

*Figure 2: Relevance of issues by number of words, 2019 dialogues.*

Another way of making use of latent issue extraction is to delve deeper into each issue by analysing the comments associated with it. This step allows for the more granular analysis of party positions by exploring what parties actually said on a particular issue.

The tool for latent issues extraction adds an additional column to the csv file containing all dialogue data to include the categorisation of each comment into the correspondent latent issue. This file can be used to filter the content related to specific issues, and also combining it with other previous filters as the ones allowing to select a specific party or time period.

## 8. Measuring party distances: transformers representations

The categorisation of text through issue extraction, both via the predefined issue extraction and the latent issue extraction, now provides a basis for the identification of party distances, that is, the substantive distance of one party position to another. The primary goal was to evaluate whether the mediation efforts had led to any convergence between parties. But it also allows for more detailed insights into which parties diverge on which issues and the behaviour of particular parties throughout the dialogue. This type of detailed analysis may be particularly useful in dynamic and complex conflict settings with multiple actors and multiple areas of conflict.

The categorisations of text via latent issue extraction and predefined issue extraction have allowed the extraction of what each party has said about each of the issues. To measure party distances, the two text categorisations were used to apply a technique that enables the evaluation of the proximity between texts. This textual distance is evaluated based on the words used and the context in which those words are used.

### 8.1. Methods

To reach an understanding of party distances, we employed Bidirectional Encoder Representations from Transformers (BERT; Devlin et al., 2019), another language model for NLP. This model is not based on a formal definition of language but is derived from a statistical understanding of how language is used. To achieve this, the model has been trained with large datasets (in this case, the English Wikipedia dump and BookCorpus (Zhu et al., 2015) a dataset of 11,038 unpublished books). During this process, it assigns a "linguistic position" for each word in an abstract multidimensional space. Words with a similar meaning are assigned positions in close proximity in this space. The model can then produce insights about the relationship between two words by measuring the distance and direction between two words in this space. For example, the distance and position of the words "London" and "UK" is the parallel to the words "Berlin" and "Germany". In this case, the representation of the words is contextual; the position of each word is also defined considering the rest of the sentence in which the word has been used (e.g., the word "pupil" has a very different meaning when talking about "eyes" or when talking about "students").

This method is applied to the dialogue dataset to assess the distance between party positions. The assumption is that the model would be able to extract party differences by measuring the distance of the language used by one party from that of another party.

In this project we used the BERT implementation of HuggingFace[8] 'bert-base-uncased' with 12-layer, 768-hidden parameters, 12-heads, and 110 million total parameters. The pre-trained model used can be found in the previous link. Texts longer than the token limit were split in order to avoid truncation. For long texts the embedding representation was obtained as the mean of the representations of its components. The distance between terms was computed as the cosine similarity, by using the Gensim library.

### 8.2. Party distances in predefined issues

We produced several graphs to illustrate party distances. In the first set of graphs, party distances are measured against an average linguistic position of all parties (Figures 3 and 4).

The top straight line represents the average position. This is not the position of a specific party but an average position of the four selected main parties. Having defined the position of each party in that "linguistic space" (calculated from the phrases used by them in the dialogues), we can also calculate in that same space what would be the average position between them, which would represent the position of consensus closest to all of them. The line of each party then represents each party's distance from the average position (the lower in the graph the farther from the average position).

The second graph (Figure 4) displays party distances against the position of a particular party, whose position has been chosen as a baseline.

---

[8] https://github.com/huggingface/transformers

Both graphs represent the average position of each party over one year of negotiations (2019). It is important to note that, at this level of abstraction and with the limited amount of data available, the graphs cannot display the fluctuation of party positions during one particular dialogue session.

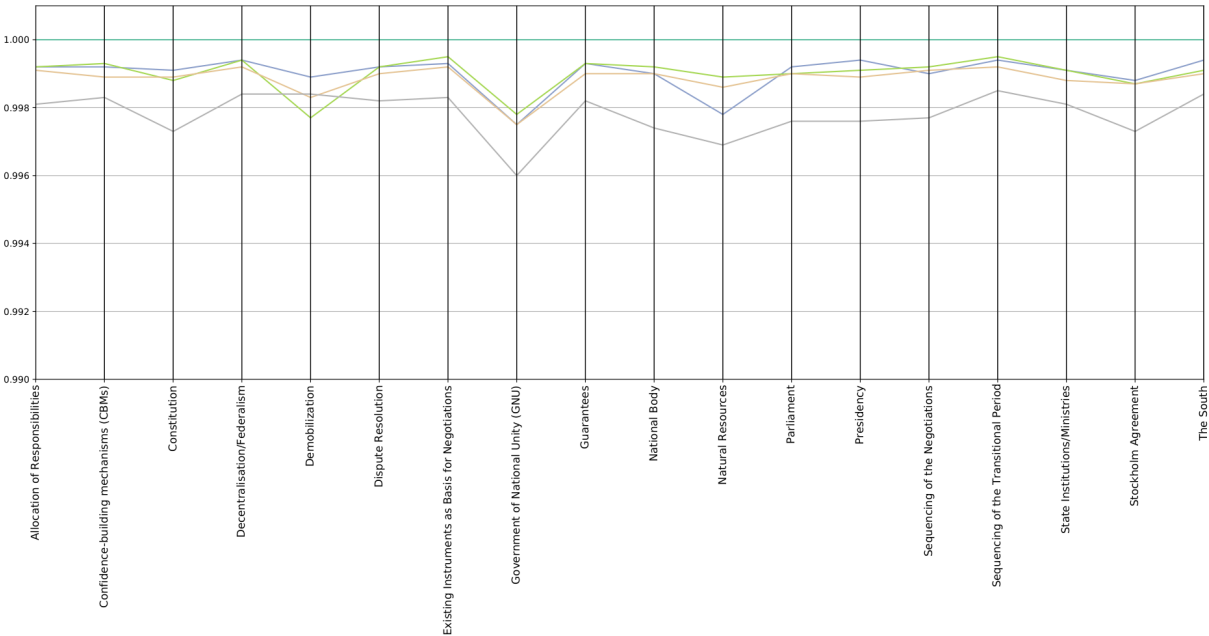

*Figure 3: Party positions measured against average party position for 2019.*

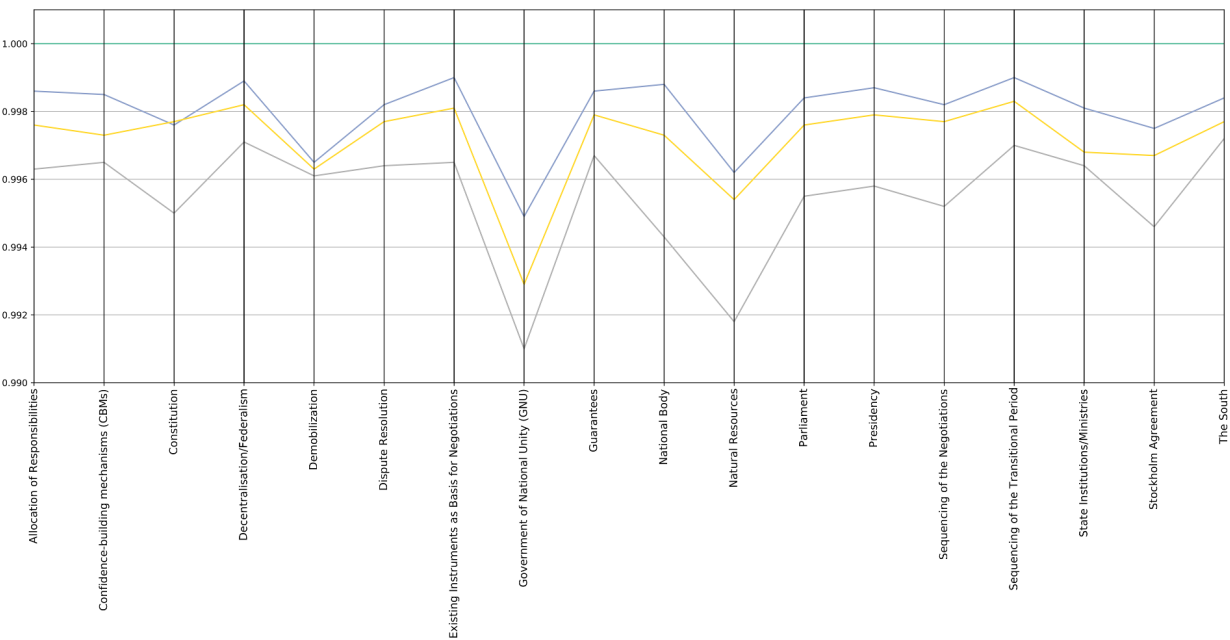

*Figure 4: Party positions measured against a baseline position of a single party for 2019.*

It is also important to clarify that in the two graphs each party represents its distance only from the reference position (average or baseline party). The fact that the lines of two parties cross does not mean that they are close to each other, it only indicates that both are at the same distance from the reference position. For the purpose of interpreting the graphs and the distances between parties it is useful to bring to mind the image of "parties in a room". If the centre of the room is the average position of all parties, then parties could be equidistant to the centre, but positioned at different corners of that room. Hence, parties that appear to be close on the graph could hold diametrically opposed political views. To inspect the specific distance between any two parties we need to refer to the next set of graphs (heatmaps), which show this more detailed information about each pair of parties.

It is important to highlight that the limited amount of data available generates a margin of uncertainty in the graphs. It is difficult to obtain a precise value of this uncertainty, but we have observed that the size of the differences between parties is in the order of magnitude of the changes observed in any party position when varying the 10% of the text of the party. Thus, the variations should be considered within the uncertainty margins. This implies that the above differences between parties should not be taken literally but should only be used as indicators to serve as a prompt for internal reflection on the mediators' own perception of the dialogues. For example, when using the above graphs mediators should focus on relative trends between parties, and not on the absolute values of the differences.[9] Generally, charts should always be analysed in conjunction with the activity chart depicted in Figure 1. In cases where the amount of information is low, there is a considerable increase in the margins of uncertainty for the distances. This means that in those cases peaks observed in these issues in the last two graphs have a much higher probability of being produced by this lack of data, than by a difference between the positions of the parties.

The most significant dips in the graphs, such as the systematic positioning of one of the parties as the most distant from the reference positions, or the increase in distance of some issues, are the clues we get from these graphs that invite a more detailed internal analysis of these matters.

Heatmaps complement the analysis of party positions shown in the graphs as they allow for a closer examination of the distance of different parties to one another and can be used for a more focused analysis of party distances on particular issues. For example, the graph depicted in Figure 4 reveals a greater distance on the "Natural Resources" issue. This would prompt a closer analysis of this issue on the basis of the heatmap to explore more closely which parties diverge. Below we can see one of these heatmaps (Figure 5).

The parties appearing in the row and column of each tile are the same as those used in the comparison. Four levels of distance are considered, going from light green meaning smaller distance (that is why the diagonal presents this colour, since it shows the position of a party compared to itself) to red, meaning larger distance.

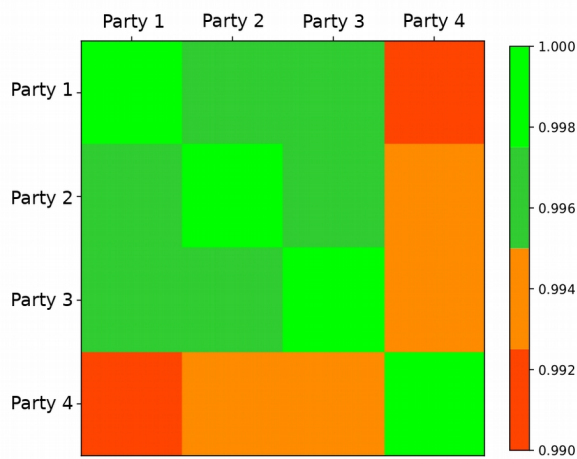
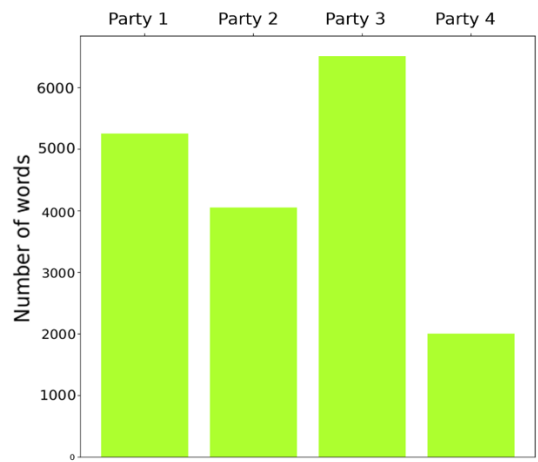

*Figure 5: Party positions between each pair of the selected parties for the issue "Natural Resources" for 2019.*

*Figure 6: Number of words of each party for the issue "Natural resources" for 2019.*

Again, as explained previously, it is important to emphasise the uncertainty in the data. To assess the viability of the data it is important to scrutinise the amount of available data not only for the whole issue but for each specific party when talking about the issue. To this end, we can make use of the party activity graph (Figure 6).

---

9     For a more detailed discussion of the values in these graphs refer to the Appendix.

Besides helping to assess the graphs, the comparison between the activity of different parties on a specific issue could prove useful when it comes to understanding whether there are parties that are dominating dialogue sessions on particular issues. Such insights could support the strategic organisation of the dialogue sessions. For example, they could prompt dialogue organisers to reconsider the allocation of speaking time; or help identify potential bottlenecks in negotiations and engage in bi-lateral dialogues if a particular issue appears to be of significant importance for a particular party.

Meetings the mediation team had with external stakeholders prompted discussions on visualisation techniques, revealing that heat maps were perceived to be particularly useful: "when we were presenting it to the envoys' office, they tended to be a lot more interested in the heat maps than the graphs. I'm not sure why. Maybe it's just because it's easier for them to understand. But there was definitely more interest in the maps." (Feedback session, 2022).

### 8.3. Party distances in latent issues

The same process was repeated with the results produced by the automatic generation of issues.

Next, we present graphs comparing the party distances on different issues. As explained in Section 5.2, in this case the issues on which the distance is evaluated have not been selected manually but generated automatically from the text. These issues are defined by a series of keywords listed in Table 1, where each one has been given a label based on its keywords and an analysis of the representative comments on each issue.

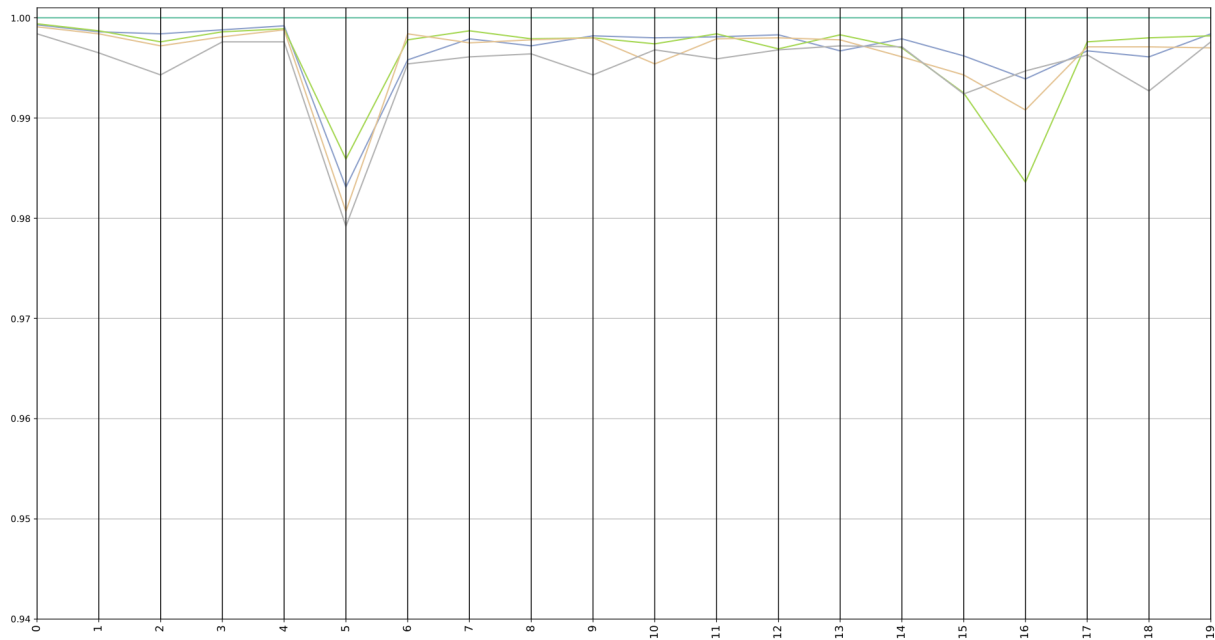

*Figure 7: Party positions measured against average party position for 2019.*

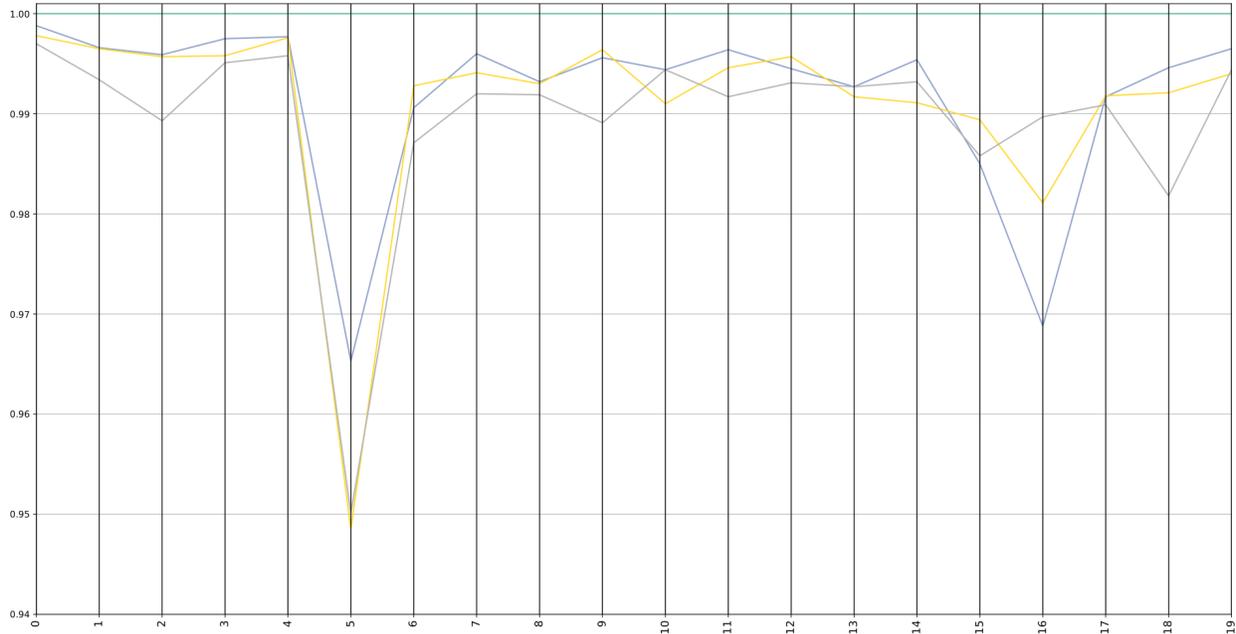

*Figure 8: Party positions measured against average position of a baseline party for 2019.*

As in the query driven case, the analysis was conducted by taking as a reference the average position of the parties and the position of a specific party. It is important to emphasise that the large dip in issue 5 is directly related to the lack of data on that issue as shown in Figure 2. For a more general discussion on these graphs refer to the previous query driven section.

Comparing both graphs with the predefined issues case it can be seen how the distances are reduced here. This may suggest that restructuring the conversations into the latent issues that emerge from the comments made by each party could make it easier to find consensus positions between the parties.

Below is an example of a heatmap (Figure 9) and its associated party activity graph. We have selected this issue from the previous graph as one of the issues where the largest differences are observed, which we observe again in this heatmap. However, we observe that in this example the amount of data is only 25% of the example in the previous heatmap. Hence the differences between party positions have larger uncertainty margins and should be considered less robust than issues that have more data.

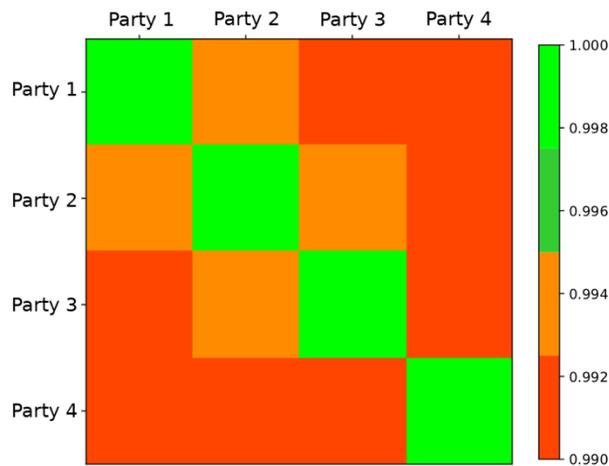

*Figure 9: Party positions between each pair of the selected parties for issue 18 for 2019*

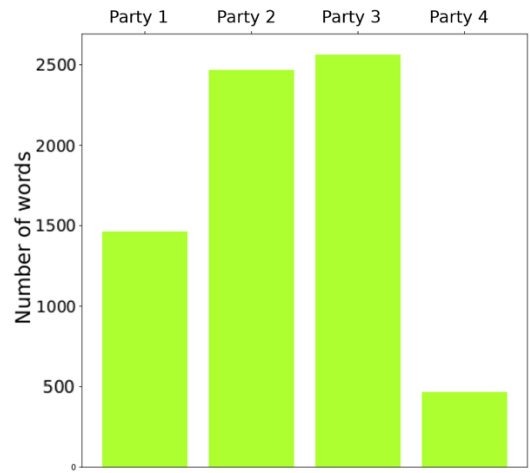

*Figure 10: Number of words of each party for issue 18 for 2019.*

## 9. Discussion

Overall, the study supports growing evidence that machine learning tools have the potential to provide meaningful insights into highly complex and dynamic mediation processes and to support consensus finding by providing tools for effective knowledge management and innovative dialogue analysis. The feedback[10] received from the mediation team was overall very positive and confirmed the potential for the application of machine learning in this particular domain. The mediation team leader commented: "the methodology as a whole seemed really interesting to people and I have been strongly encouraging other organisations to do the same, not just in Yemen, but elsewhere as well" (Feedback session 2022).

### 9.1. Knowledge Extraction and Management

At the most basic level, machine learning tools can effectively support mediators in extracting and managing knowledge accrued over several years of peace dialogue. Often, as in the case at hand, valuable data may be available but is too complex and too vast to be assessed without technological support. The tools developed in this project have demonstrated that machine learning can bring significant efficiency to mediation by organising complex data and making it more easily accessible. The Predefined Issue Extraction Tool, for example, has significantly eased the process of organising large amounts of dialogue text into predefined issues – a task that previously took several researchers weeks to accomplish. While machine learning driven categorisation may not yet be as precise as a human-driven one, the tool still allowed the organisation to revise and update their "Comprehensive Analysis" more rapidly and conveniently. The tool also significantly improved the accessibility of available data. Presenting and organising data outputs in a CSV file allows mediators to perform targeted searches by filtering information according to different variables. For example, mediators can now quickly obtain information on what a particular party had to say on a particular issue at any given time or track how the position of a party on an issue has changed over time. This could then be used to further investigate why shifts in a party's position may have occurred and to hold parties accountable to particular positions or to concentrate work on particularly controversial issues.

### 9.2. Mediation Process Management

Beyond knowledge extraction and management, the feedback from the mediating team emphasises how the tools have helped them to better grasp conflict dynamics and party positions and to adapt their dialogue strategy. A deep understanding of a conflict situation, the key conflict areas and how they may be interrelated, its actors and their interests, the relationship between different actors, and their potential openness to finding alternative solutions is key to devising an effective dialogue strategy (Amon et al., 2018). Machine learning tools may be particularly useful in complex and lengthy mediation processes in which viewpoints may otherwise calcify and prevent consensus finding by helping to challenge stereotypes and prejudices that may have built up over time (see Hirblinger 2022, p. 20).

Machine learning tools can offer new perspectives on dialogue dynamics and so provide evidence to help mediators adapt their dialogue strategies:

- The Latent Issue Extraction tool can be used to identify potentially overlooked but significant dialogue issues, to detect if new issues have emerged and others receded or to detect which parties are closer to the average positions and which parties are further away.
- The Party Distances Tool can be used to identify "Zones of Possible Agreement" or better understand which issues provoke the most disagreement between particular parties. This insight can then be used to focus on issues where party positions appear to be converging and create momentum for successful dialogue. Alternatively, mediators could concentrate their efforts on consensus-finding on issues identified as acting as bottlenecks.
- The party activity graph can help identify if particular parties are dominating dialogue sessions on specific issues.

---

10    Through our participatory and co-design methodology we received feedback throughout the tool development process. We sought more detailed reflections from the mediation team on the process, on potential use cases and strengths and weaknesses of the tools during a dedicated feedback session in April 2022.

Such insights can provide the basis to fine-tune dialogue proceedings by considering allocation of speaking time or by engaging in bi-lateral negotiations with a particular party. As a member of the mediation team explained: "You can imagine saying [to the parties]: it looks to us that these are the following three areas where agreement is more likely to be difficult. Just to warn you that you should perhaps think more about these issues rather than others or invest more of your time during the preparation period to try to find solutions to these particular issues [...]. Strategic dialogue focus can also be supported by using the analysis of overlapping issues to identify connections between different issues and to find ways of addressing them. In particular, members of the mediation team felt that "the data can be used also to set the agenda for the meetings as well and to determine the format for some of the negotiations [...] So, if there's a negotiation session that's about to happen, then the data could be used also to prepare the agenda, not just to prepare the participant, to prepare the negotiators and mediators, but also the format of the negotiation and the order in which things are going to be discussed as well." (Feedback session 2022).

The mediating team also commented on the likely broader impact of the methodology on their relationships with external stakeholders, including funders and making them more accountable: "[B]ut I think it's important that as soon as we presented it to the mediation supporting unit at the Foreign Office, the immediate reaction was this is potentially a great tool for measuring whether or not we're having an impact." Especially in complex and protracted mediation scenarios, progress is often notoriously hard to track:

- The Party Distances Tool can be used to examine whether party positions have converged over time, stagnated or diverged further, and to identify the impact of external factors on the dialogue sessions by analysing how party positions have shifted at particular points of the dialogue proceedings.;
- The issue activity graphs can be used to identify if the organisation of the dialogues is congruent with the issues that were prioritised in the dialogue activity of the parties, and how party activity on each issue fluctuates over the years..

The analysis of party distances from 2018 to 2019 for both the predefined issues and the latent issues shows that, overall, party positions appear to be converging, although this should be confirmed via a detailed analysis of dialogue transcripts. Also, when analysing the number of words per issue graphs, the data shows a significant increase in dialogue activity from 2018 to 2019. Such analyses can prove useful for a meta-level assessment of the overall trajectory of peace negotiations and the impact of the mediators' efforts on consensus finding. Indeed, the mediation team used the data analysis emanating from this project in exchanges with several external stakeholders. In particular, it was emphasised that the tools offered a more systematic form of assessing the impact of a mediation intervention and as a way of increasing the accountability of mediators (Feedback session 2022).

Another aspect mentioned in the mediation team's feedback was that the tool can be used to "objectify" the course of the proceedings. Peace dialogues often rely on analyses of people, who may introduce biases in the interpretation of particular conflict dynamics and developments. This is particularly important for protracted negotiations, where new members of negotiation teams may need such information. As the mediation team commented: "it's not enough anymore for people just to rely on bilateral meetings and their impressions. But everything should be analysed afterwards through a more scientific method" (Feedback session 2022).

Of course, the analysis produced by digital tools should not be taken as the ground truth regarding the dialogue participants' positions - the tools will merely reflect what dialogue participants have said during the dialogues. Dialogue participants constantly make strategic choices about their voiced standpoints - they may hide real interests and negotiation positions at times, or resort to posturing and other negotiation tactics. The data analysis provided by machine-learning tools will have to be interpreted accordingly.

## 9.3. Challenges and limitations

Despite the potential that these tools may bring to mediation processes, it is important to highlight further challenges and limitations relating to data, machine-learning-based analysis techniques, and the way that such tools are being used in practice.

### 9.3.1. Data quantity and quality

One of the key limitations of the project has been the limited amount of data available for analysis. This limitation in terms of the volume of the data is likely to affect data analysis as it has the potential to skew results and to misguide its interpretation. This is particularly evident in the case of measuring party distances. Here, limited data prevents a more detailed analysis of party positions over time and can obscure party differences and skew results, potentially leading to misguided interpretation. This problem is exacerbated when more detailed analysis is sought for shorter timeframes, as the amount of data will be further reduced. Further, data analysis for this project was conducted on "rough notes" that were translated from Arabic into English, rather than verbatim transcripts of dialogue sessions. While we are confident that the data still reflects the key features of the deliberations, verbatim transcripts will reduce the risk of bias and error introduced by note-taking and translation. As mentioned, mediators will need to be conscious of these limitations when using the tool and merely use its outcomes as a prompt to corroborate them via deeper qualitative analyses of original dialogue text.

### 9.3.2. Refining the tools

The project also highlighted persisting challenges for techniques in identifying nuanced vocabularies and their context and dealing with word play and ambiguity (Höne, 2019, p. 12). These were particularly evident when aiming to distinguish between closely related issues that address overlapping subject matters. In the case at hand, for example, issue extraction tools struggled to separate related but independent issues dealing with institutional matters, such as the issue of "National Body" or "Government of National Unity" or with procedural questions of the transition, such as "Sequencing of Negotiations" and "Sequencing of the Transitional Period". However, more work will need to be done to improve precision when dealing with linguistically and substantively closely related matters.

### 9.3.3. Methodology and trust in machine learning

Complex and sensitive mediation processes remain a human-centred trade for which machine learning tools can only offer a degree of analytical support. At present, ensuring "meaningful human control" (Höne, 2019, p. 11) over what Hirblinger (2022) refers to as 'hybrid intelligence peacemaking systems' is best achieved when mediators are actively involved as collaborators throughout the development process. Continuing to follow a participatory, co-creation methodology will also help the mediation team to develop further their understanding of the kind of problems that machine learning tools can provide assistance with. It will also contribute to robust, ethical and context-sensitive tool development that can substantially contribute to the mediation teams' understanding of the conflict.

It was clear from the feedback from the mediation team that their trust in the tools developed had its foundations in the project's participatory, co-creation methodology, which enabled them to query how the tools work, seek clarification about the meaning of the results they generated and thereby retain the necessary oversight of the mediation process (Hirblinger, 2022). As part of ongoing collaboration with our collaborators, we will focus on how to create and present explanations of the analyses produced by the tools that will satisfy these requirements. For example, we will investigate the design of dashboards that can present outputs in more accessible ways and thus help guide responsible and informed interpretation (Bell et al., 2021).

Finally, there is the question of whether trust in machine learning tools and meaningful human control of mediation processes can be achieved and sustained without the methodological scaffolding used in this project. This might seem to be a necessary advance if machine learning tools are to achieve wider application, both within this domain and more widely. We are not at present able to answer this question with confidence, however, it is clear that responsible and effective human-centred use of machine learning tools means future mediators will need data literacy skills, a grasp of the technological underpinnings of such tools, along with a methodological framework (i.e., "from the data to the graphs and back to the data") that will enable them to critically analyse outcomes, as well as to explain and justify the decisions that follow from these analyses.

## 10. Concluding remarks

While data analytics has been employed in broader humanitarian and peacebuilding contexts, e.g., for the purpose of conflict analysis, early warning, prediction of conflict, such tools have not been extensively used in the context of mediation efforts. The application of machine learning in this case study has shown that these tools can play a significant role in forming comprehensive conflict analyses and informing mediation strategies. Such tools become particularly pertinent in the context of the emergence of more dynamic and complex conflicts. This study demonstrated that machine learning tools can effectively support mediators: (1) in managing knowledge by analysing large amounts of unstructured data; and (2) by providing them with new analytical tools that may lead to new perspectives on a conflict scenario.

The project also emphasised that machine-learning tools cannot replace human analysis, particularly in highly sensitive contexts such as conflict mediation. Meaningful and responsible development and deployment of machine learning tools requires an interdisciplinary and participatory, co-creation methodology to help develop an understanding among users of machine learning's capabilities and limitations and to help data scientists to create context-sensitive, targeted, effective, trustworthy and ethical tools.

The introduction of new tools into work practices almost inevitably requires adaptation on the part of their users; this, in turn, leads to the need to adapt the tools, as users gain familiarity with them, discover their strengths and weaknesses (Hartswood et al., 2002a; 2002b; Bodker & Kyng, 2018). This, we argue, applies particularly to machine learning tools, such as the ones described here, that are intended to assist – and not substitute for – human interpretation of data. The feedback from the mediation team was very positive and confirmed the potential for the application of machine learning in this particular domain. However, owing to the pausing of the peace negotiations, the evidence is currently limited to the mediation team's use of the analysis generated during the project. Hence, we are continuing to work with the mediation team in order to learn from their experiences of using the tools and thereby help to evolve them in ways that are productive for conflict mediation.

## 11. Funding

This research was funded by International IDEA under grant number TE-30 (FM-PR-01).

## 12. Acknowledgements

We would like to thank the team at International IDEA for their contribution to the project.

An earlier version of this paper was presented at the Data for Policy Conference, September, 2021. It is available on arXiv at https://arxiv.org/pdf/2108.11942.pdf

## 13. Data Availability

The data used in this research was provided by International IDEA. Requests to access the data should be directed to them.

## 14. Competing Interests

The authors declare that they have no competing interests.

## 15. Author Contributions

Miguel Arana-Catania: Conceptualization, Investigation, Methodology, Visualization, Writing – original draft, Writing – review & editing, Data curation, Formal analysis, Software, Validation

Felix Van Lier: Conceptualization, Investigation, Methodology, Visualization, Writing – original draft, Writing – review & editing, Data curation, Funding acquisition, Resources, Validation

Rob Procter: Conceptualization, Investigation, Methodology, Visualization, Writing – original draft, Writing – review & editing, Project administration, Supervision, Funding acquisition

## Appendix A: Structure of the word representation space

In this project we worked with numerical representations of text dialogues, and it is therefore important to take into account the structure of these representations and how this structure can affect the analysis of these texts. We have applied different models to represent texts, but we will limit ourselves in this appendix to studying the case of GloVe, although the conclusions can be extended to other models such as BERT.

As mentioned previously, we have used the 'glove.6B' pre-trained embeddings of GloVe. This model has a vocabulary of 400,000 terms, each of which is represented by a vector with 300 components. If we analyse how these vectors are distributed in this 300-dimensional space, we can see that they are distributed anisotropically. In Figure A.1 we represent for each of the vocabulary words which is its largest component; in other words, towards which of these 300 directions each of the vectors is mainly oriented. As we can see, there are directions with a higher density of vectors than others. Figure A.2 shows the same information as a histogram, confirming the prevalence of some directions over others and therefore the anisotropy of the space of representations.

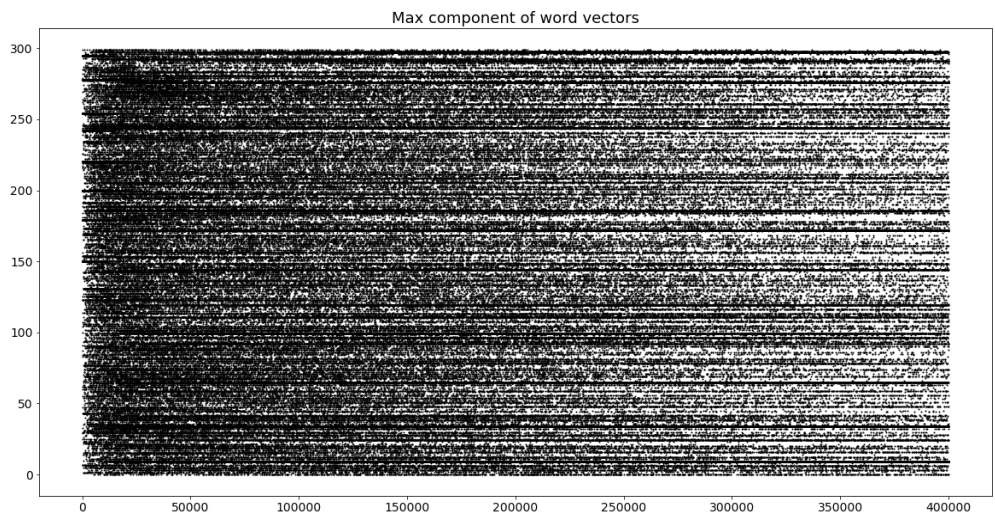

*Figure A.1*

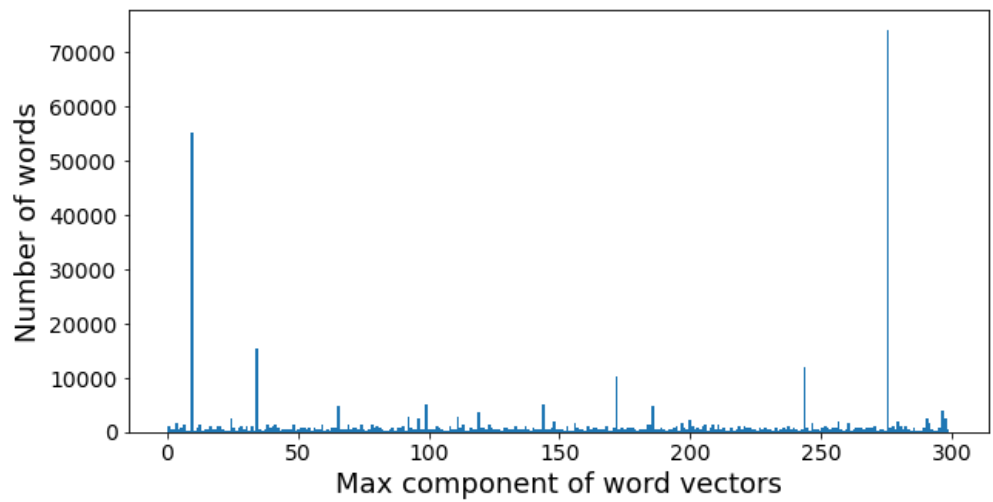

*Figure A.2.*

In the following, we will show the effects of such a structure when operating on long texts. To show the universality of this analysis, we will take as an example of analysis two corpora which are very different from each other and also with regard to the texts used in this project: Corpus 1 = the text of 'The Wisdom of Father Brown' by G. K. Chesterton; Corpus 2 = the text of 'Leaves of Grass' by Walt Whitman.

In Figure A.3 we plot in the top row the value of the largest and smallest components of each of the first 5000 words of Corpus 1 (we include the smallest to take into account all possible orientations of the vectors). In the bottom row we represent the same quantities, but in this case for the average vectors up to each term; for the term number n we take the average of the first n vectors. As we can see in this row, although at the beginning the extreme values oscillate, when a sufficient number of words are considered, they converge towards a value with respect to which they remain approximately stable. In Figure A.4 we can see the same behaviour in the case of Corpus 2.

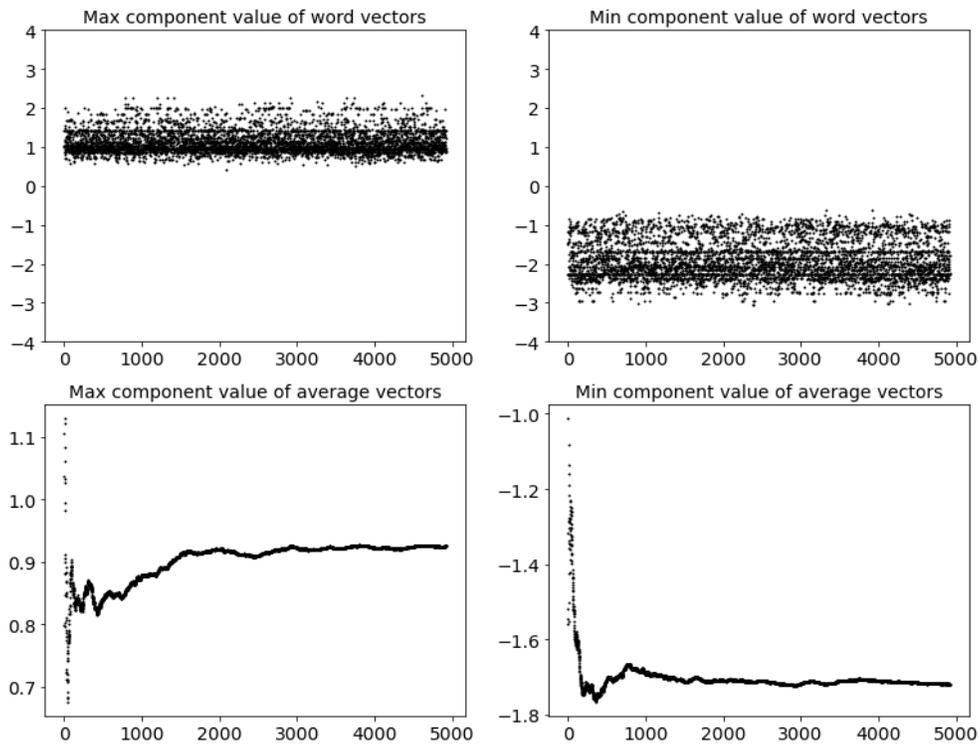

*Figure A.3.*

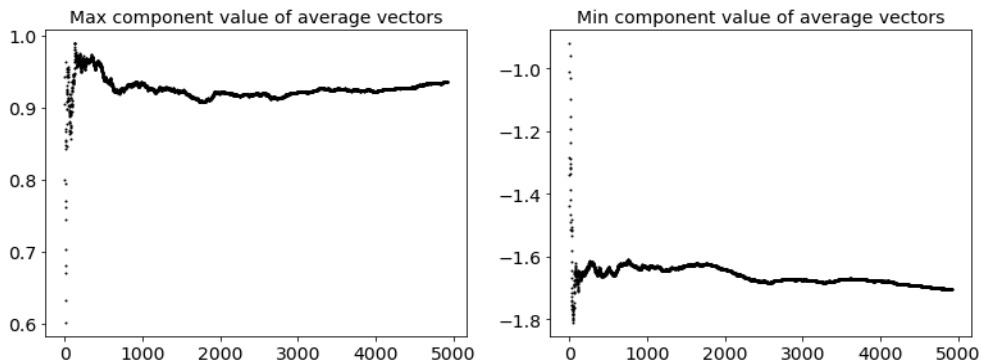

*Figure A.4.*

In Figure A.5 we observe again the effect of averaging over the terms, but in this case considering all the components, not only the extreme ones. The top row represents Corpus 1 and the bottom Corpus 2. Again, we observe the effect of convergence due to the anisotropy of the space of representations as we average over more vectors, and the large text similarity between the two corpora in the last column.

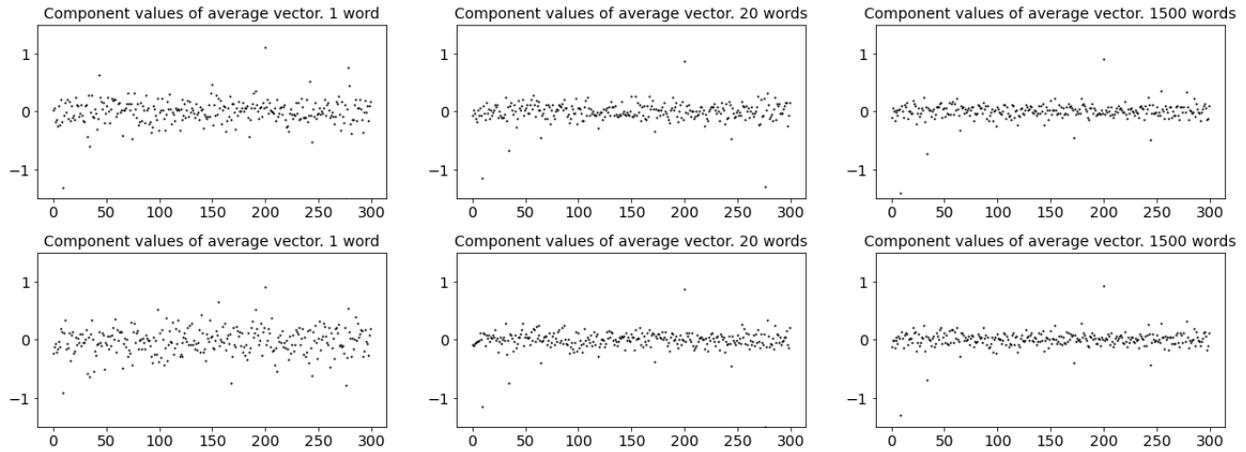

Figure A.5.

In Figure A.6 we plot the cosine similarity between the average vectors of each corpus. For the term n we compute the average vector of the first n terms in each corpus, and then we compute the cosine similarity between both resulting vectors. Although the cosine similarity measure already performs a normalisation effect, by considering only the direction of the vectors, and not their magnitude, we see here how considering sufficiently long texts has a convergence effect with respect to the values of the cosine similarity. The range of possible values will therefore reduce and converge to the unit as the amount of text considered increases. This effect must be taken into account when assessing the differences between comparisons, and especially when comparing texts of different sizes.

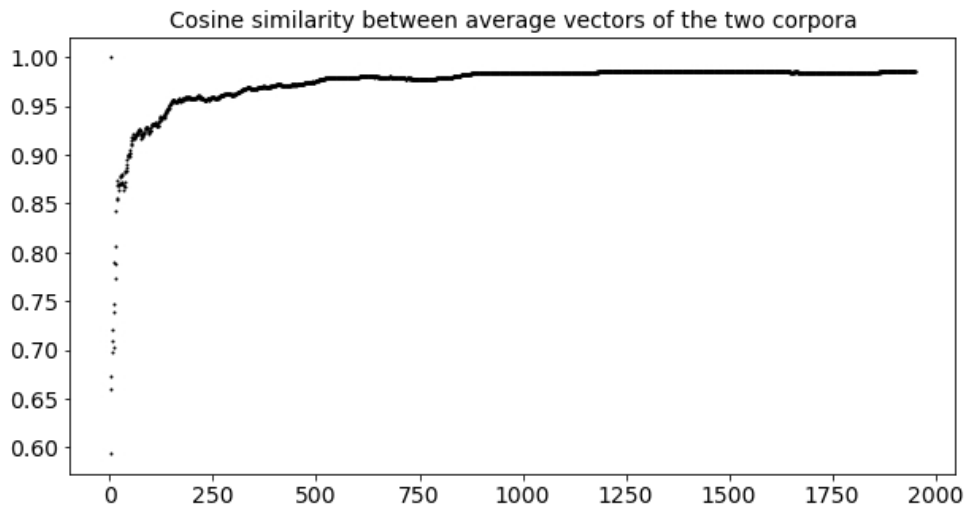

Figure A.6.